\documentclass[sigconf]{acmart}
\usepackage{xcolor}
\usepackage{tikz}
\usepackage{soul}
\usepackage{url}
\usepackage{multirow}
\usepackage{bm}
\usepackage{subfigure} 
\usepackage{caption}
\usepackage{graphicx}
\usepackage{float} 
\usepackage[pagewise,switch]{lineno}
\usepackage{makecell}
\usepackage{enumitem}
\usepackage[linesnumbered,ruled,vlined]{algorithm2e}
\usepackage{fancyhdr}
\usepackage{lipsum} 
\usepackage{multicol} 
\usepackage{hyperref} 
\usepackage{balance}

\newcommand*{\expect}[2][]{\ensuremath{\mathbf{E}_{#1} \left[ #2 \right] }} 

\newcommand*{\ptrue}{\ensuremath{\bm{p}}}
\newcommand*{\causalvar}{\theta_\rightarrow}
\newcommand*{\antivar}{\theta_\leftarrow}
\newcommand*{\pmodel}{\ensuremath{\bm{p}_{\causalvar}}}

\newcommand{\real}{\mathbf{R}}
\newcommand*{\ptransfer}{\ptrue^*}

\newcommand{\logpartition}{A}
\newcommand{\alpin}{\alpha}

\newcommand*{\smoothness}{B}

\newtheorem{proposition}{Proposition}

\newcommand{\cL}{{\mathcal L}}
\newcommand{\KL}{D_{\mathrm{KL}}}
\newcommand{\inv}[1]{\frac{1}{#1}}
\def\vs{{\bm{s}}}
\def\vc{{\bm{c}}}
\def\evs{{s}}
\def\evp{{p}}
\DeclareMathOperator{\assign}{\leftarrow}

\copyrightyear{2023}
\acmYear{2023}
\setcopyright{acmlicensed}\acmConference[CIKM '23]{Proceedings of the 32nd
ACM International Conference on Information and Knowledge
Management}{October 21--25, 2023}{Birmingham, United Kingdom}
\acmBooktitle{Proceedings of the 32nd ACM International Conference on
Information and Knowledge Management (CIKM '23), October 21--25, 2023,
Birmingham, United Kingdom}
\acmPrice{15.00}
\acmDOI{10.1145/3583780.3614774}
\acmISBN{979-8-4007-0124-5/23/10}

\begin{document}


\title{Adaptation Speed Analysis for Fairness-aware Causal Models}
\author{Yujie Lin}
\affiliation{%
  \institution{School of New Media and Communication, Tianjin University}
  \city{Tianjin}
  \country{China}}
\email{linyujie_22@tju.edu.cn}
\begin{abstract}
    For example, in machine translation tasks, to achieve bidirectional translation between two languages, the source corpus is often used as the target corpus, which involves the training of two models with opposite directions. The question of which one can adapt most quickly to a domain shift is of significant importance in many fields. Specifically, consider an original distribution $\bm{p}$ that changes due to an unknown intervention, resulting in a modified distribution $\bm{p^*}$. In aligning $\bm{p}$ with $\bm{p^*}$, several factors can affect the adaptation rate, including the causal dependencies between variables in $\bm{p}$. In real-life scenarios, however, we have to consider the fairness of the training process, and it is particularly crucial to involve a sensitive variable (bias) present between a cause and an effect variable. To explore this scenario, we examine a simple structural causal model (SCM) with a cause-bias-effect structure, where variable A acts as a sensitive variable between cause (X) and effect (Y).  The two models respectively exhibit consistent and contrary cause-effect directions in the cause-bias-effect SCM. After conducting unknown interventions on variables within the SCM, we can simulate some kinds of domain shifts for analysis. We then compare the adaptation speeds of two models across four shift scenarios. Additionally, we prove the connection between the adaptation speeds of the two models across all interventions. 
    \label{abstract}
\end{abstract}

\author{Chen Zhao}
\affiliation{%
  \institution{Department of Computer Science, Baylor University}
  \city{Waco, Texas }
  \country{USA}}
\email{chen_zhao@baylor.edu}

\author{Minglai Shao}
\authornote{Corresponding author}
\affiliation{%
  \institution{School of New Media and Communication, Tianjin University}
  \city{Tianjin}
  \country{China}}
\email{shaoml@tju.edu.cn}

\author{Xujiang Zhao}
\affiliation{%
  \institution{NEC Lab}
  \city{Princeton, New Jersey}
  \country{USA}}
\email{zhaoxuj32@gmail.com}

\author{Haifeng Chen}
\affiliation{%
  \institution{NEC Lab}
  \city{Princeton, New Jersey}
  \country{USA}}
\email{haifeng@nec-labs.com}

\renewcommand{\shortauthors}{Yujie Lin, Chen Zhao, Minglai Shao, Xujiang Zhao, \& Haifeng Chen}


\maketitle

\section{Introduction}
\label{sec:intro}
\begin{figure}
    \centering   \includegraphics[width=\columnwidth]{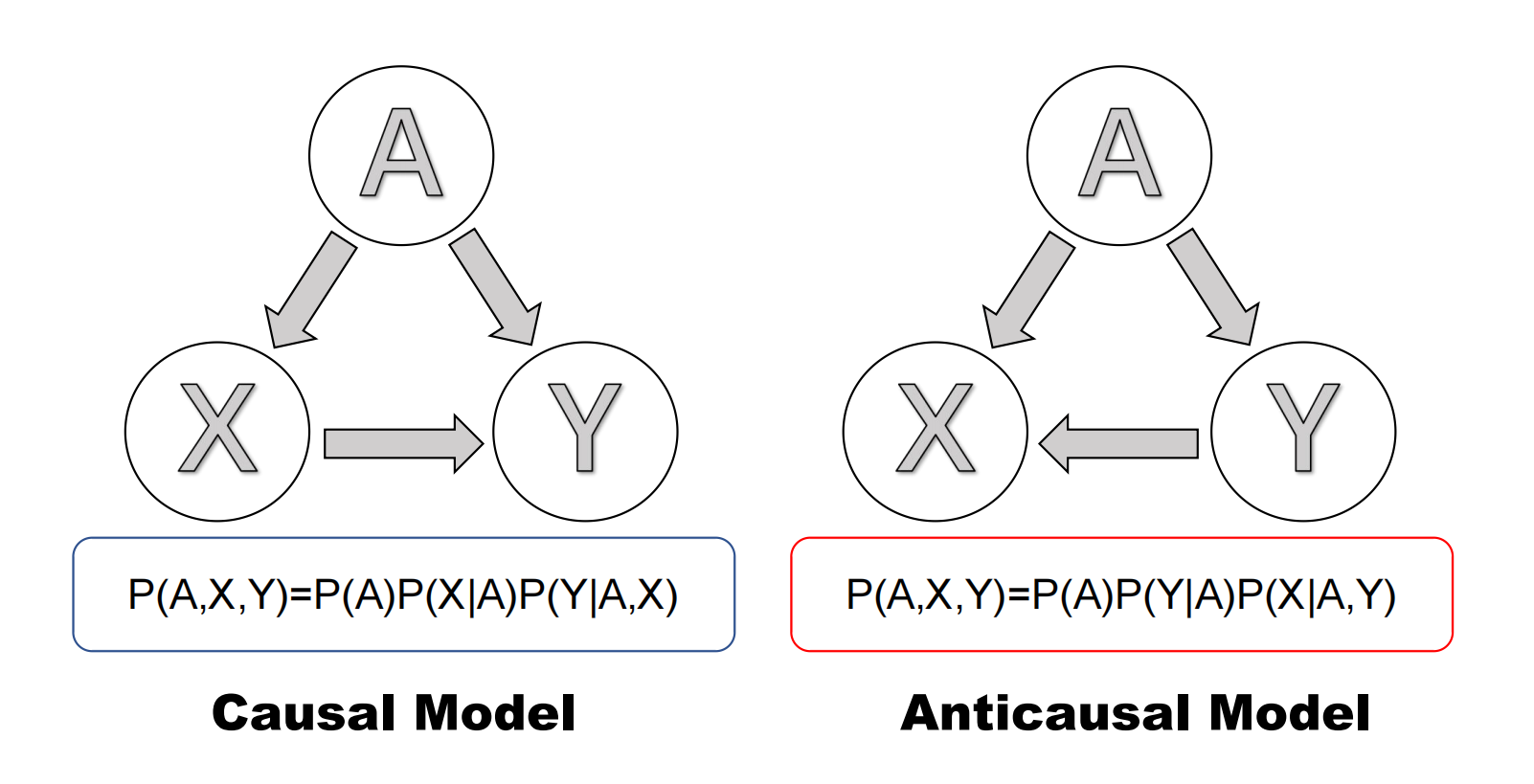}
    \caption{Two fairness-aware training models. The causal model has the same $X\to Y$ direction of the cause-bias-effect SCM, while the anti-causal model keeps the opposite direction. }
    \label{fig:two_model}
    \end{figure}

The widespread adoption of machine learning systems, particularly in decision-critical domains like criminal sentencing and bank loans, has raised concerns regarding the fairness implications~\cite{gohar2023survey}. AI systems are increasingly employed in sensitive contexts where they make significant and life-altering decisions~\cite{mahoney2020ai,trewin2019considerations,zhao2021fairness,zhao2019rank,zhao2021fairnessphd,zhao2020unfairness}. A sensitive feature is defined as an attribute that contains protected information about individuals or groups within a dataset. This information may encompass characteristics such as race, gender, religion, or socioeconomic status, which are safeguarded by ethical considerations, legal regulations, or societal norms~\cite{caton2020fairness}. Machine learning models have the potential to inadvertently acquire discriminatory patterns if sensitive variables exhibit spurious correlations with the target variable or predictive outcomes~\cite{pessach2022review,oneto2020fairness}. Consequently, this can lead to biased decisions or unfair predictions that disproportionately impact specific individuals or groups. Therefore, it is crucial to ensure that these decisions do not reflect discriminatory behavior towards particular groups or populations~\cite{mehrabi2021survey}.

Causal models have been widely applied in machine learning to address issues related to model fairness. Structural Causal Models (SCM) ~\cite{2001Causality} provide a means of explaining machine learning model predictions. Analyzing causal graphs and paths helps understand how the model's predictions for different groups are formed, thereby identifying and addressing potential unfair factors.
A simple SCM is the model $X \to Y$ where X is the cause and Y is the effect, where it indicates that X determines Y. 
Modern machine learning methods encounter surprising failures when the test distribution differs from the training distribution, which is commonly known as \textit{domain shift}~\cite{elsahar2019annotate}. Although many domain adaptation~\cite{wang2018deep,chu2018survey,saunders2022domain} and domain generalization~\cite{zhou2022domain,li2018learning,muandet2013domain} methods have been proposed in recent years to mitigate the problem of domain shift, in reality, we often need to achieve the best possible results and for the convenience of training, we still need to retrain the data in the new domain. In the process of relearning from the original distribution to the new distribution, the relative speed of adaptation between the causal model and the anti-causal model may differ. Considering this SCM ($X \to Y$), the previous work~\cite{2020arXiv200509136L} analyzed the adaptation speed of two models trained in the direction of the causal dependency (i.e., from X to Y) and in the reverse direction of the causal dependency (i.e., from Y to X) concerning domain shift. While they have obtained some promising findings, the analysis of adaptation speed did not take into account fairness considerations (\textit{i.e.,} sensitive variables), which are crucial in real-life scenarios. There are currently theoretical and experimental gaps in the analysis of this aspect, which is precisely what our work aims to address.

To comprehensively analyze the aforementioned issue, it is necessary to consider the convergence speed of the two models under different domain shifts. To achieve this, we employed interventions on various variables to simulate different domain shifts. In our analysis, three variables need to be considered, namely: bias (or sensitive variable) $\bm A$, cause $\bm X$, and effect $\bm Y$. We will perform interventions on one or more variables among these three. Specifically, we consider four scenarios, including interventions on bias and cause separately, interventions on both bias and cause, and interventions on the effect.

Having identified the four types of domain shift that we need to consider, now let us provide a more formal definition. The three variables (bias: A, cause: X, effect: Y) related to fairness can form a cause-bias-effect structural causal model $(A\to X$ and $A,X\to Y)$. We consider two training models with sensitive variables: one that has the consistent cause-effect direction of the cause-bias-effect SCM, and the other that has the opposite direction (Fig.~\ref{fig:two_model}). Domain shift can be described as the transition from the initial joint distribution $\bm p(A,X,Y)$ in the training set to the joint distribution $\bm p^*(A,X,Y)$. During the alignment process of distribution $\bm p$ to $\bm p^*$, we extract several samples from $\bm p^*$ in each round and utilize them to calculate the loss value. This approach aims to gradually converge the distribution towards $\bm p^*$. Through experimentation and formula derivation, we were surprised to discover that interventions on bias causing domain shift only affect the absolute convergence speed of the two models, but do not impact the relative speed between the causal model and the anti-causal model. In other words, the faster adaptation to the new distribution by either the causal model or the anti-causal model is solely determined by interventions on other variables. We observed that in only one specific scenario of domain shift, the two models maintain a consistent relative speed relationship. When interventions are applied to the cause variable, the causal model consistently exhibits a speed advantage. However, when interventions are applied to the effect variable, the convergence speed of the causal model is faster only under certain conditions. Otherwise, the anti-causal model has a faster adaptation speed. In conclusion, our \textbf{contributions} can be summarized as follows:
\begin{itemize}[leftmargin=*]
\item 
Our work provides a significant contribution to the understanding of the adaptation speed of causal models in scenarios involving sensitive variables. Our work is the first of its kind and offers valuable information from the perspective of fairness.
\item We considered domain shifts caused by four types of interventions on the data distribution. And we analyze the relationship between the adaptation speed of two different causal models and provide insights into this topic, using both synthetic and real data.  The results can guide the development of more efficient algorithms for learning causal relationships and improving our understanding of complex systems.
\item  We conduct theoretical analyses for each of the four cases. The proof results are consistent with the conclusions drawn from the experiments, ensuring the rationality and correctness of the findings.
\end{itemize}

\section{Related Work}
\label{sec: Related Work}
\textbf{Domain shift}, which is also known as dataset shift or covariate shift, has received considerable attention in recent years~\cite{stacke2020measuring,luo2019taking,zhang2021adaptive}. It is a common problem in machine learning where the training and testing data are drawn from different distributions. This difference in distribution can cause a significant drop in the performance of the learned model on the testing data. Domain shift can occur due to various reasons, such as differences in data collection procedures, environmental changes, or task-specific variations. To address domain shift, several approaches have been proposed, including re-weighting \cite{sugiyama2007covariate}, importance sampling \cite{ganin2015unsupervised}, and transfer learning \cite{pan2010survey}. These methods aim to either adjust the training data distribution to match the testing data distribution or to learn a mapping from the source domain to the target domain. Some deep learning-based methods, such as deep neural networks, have also shown promising results in domain adaptation tasks \cite{long2015learning}. These methods learn a representation of the data that is invariant to the domain shift, enabling the model to generalize to new domains.

\textbf{Causal fairness} aims to prevent machine learning models from perpetuating or amplifying unfairness inherent in the underlying data-generating process. Numerous studies have proposed methods to achieve causal fairness, including the utilization of causal inference to identify and adjust for confounding variables that may introduce bias in predictions. For instance, previous work proposed a method based on counterfactual regression, allowing for the estimation of causal effects between protected attributes (e.g., gender or race) and the desired outcomes \cite{Kusner2017}. Other researchers have advocated for the use of causal graph structures to represent variable relationships and ensure the model adheres to specific causal consistency conditions \cite{Peters2017}.  Additionally, the application of causal models has been explored in reasoning about counterfactual fairness \cite{Kilbertus2017}.

\section{Background}
\label{sec:method}

In this section, we first provide an overview of Structural Causal Models (SCM) and interventions that can cause domain shifts.
We then introduce the formalism used in the analysis, which includes observations, interventions, models, and adaptation, as presented in the analysis~\cite {2020arXiv200509136L}. And the formalism was first proposed by the work~\cite{2020A}.
\par
\subsection{Structual Causal Model (SCM)} Structural causal models (SCMs) are widely used in causal inference to model the causal relationships among variables. An SCM consists of a directed acyclic graph (DAG) and a set of structural equations that define the causal relationships among the variables in the graph \cite{pearl2009causality,spirtes2000causation,pearl2018book}. The structural equation for an endogenous variable $X_i$ can be expressed as follows:

\begin{equation}
X_i = f_i(\textbf{Pa}_i, U_i)
\end{equation}
where $\textbf{Pa}_i$ denotes the parent set of $X_i$ in the graph, and $U_i$ denotes the set of exogenous variables that directly affect $X_i$. The function $f_i$ represents the causal relationship between the parent variables and $X_i$.
SCMs are used to estimate causal effects and test causal hypotheses. By including sensitive variables in the graph and modeling their causal relationships with other variables, SCMs can adjust for sensitive and produce unbiased estimates of causal effects \cite{hernan2018causal}.

\par
\textbf{Interventions on SCMs}
 involve changing the value of a variable to a specified value. This can be represented mathematically using the do-operator, denoted by $\mathbf{do}(X=x)$. The do-operator separates the effect of an intervention from the effect of other variables in the system. For example, if we want to investigate the effect of drug treatment on a disease outcome, we might use the do-operator to set the value of the treatment variable to "treated" and observe the effect on the outcome variable. In the following narrative, we will use $\bm p^*$ to represent this modified distribution, such as 
\begin{equation}
    \bm p^*(a,x,y)=\bm p(a,x,y|\mathbf{do}(x=t)).
    \end{equation}

By controlling one or several variables in this way, we simulate domain shifts under different scenarios. And $\bm p^*$ is the outcome after domain shifts.

\subsection{Reference and Transfer Distributions} We obtain the initial reference distribution $\bm{p}$ by sampling the triad $(A,X,Y)$ from a structural causal model (SCM) constructed as follows: $A$ is a bias, $X$ is the cause, and $Y$ is the effect. The SCM is defined by the following two equations: $A \to X$ and $A,X \to Y$. Now, we perturb the distribution $\bm{p}$ by performing interventions on certain variables to obtain the following transfer distribution $\bm{p^*}$.
\begin{table}[!h]
  \centering
  \caption{\textbf{Tranfer distributions of different interventions.}}
  \begin{tabular}{ | c | c | }
    \hline
     Intervention & Transfer Distribution\\ 
     \hline 
     Bias & $\bm{p}^*(a,x,y) = \bm{p}^*(a) \bm{p}(x|a) \bm{p}(y|a,x)$\\
    \hline   
     Cause & $\bm{p}^*(a,x,y) = \bm{p}(a) \bm{p}^*(x) \bm{p}(y|a,x)$\\
    \hline   
     Bias and Cause & $\bm{p}^*(a,x,y) = \bm{p}^*(a) \bm{p}^*(x) \bm{p}(y|a,x)$\\
     \hline   
     Effect & $\bm{p}^*(a,x,y) = \bm{p}(a) \bm{p}(x|a) \bm{p}^*(y)$\\
    \hline
  \end{tabular}
   
    \label{table:Tranfer distribution}
 
\end{table}

If the intervention is on the bias, we sample $A$ from a different marginal distribution, while $X$ and $Y$ are sampled from the reference conditional distribution.
If the intervention is on the cause, $A$ is sampled from the reference marginal distribution, $X$ is sampled from a different marginal distribution independently of $A$, and $Y$ is sampled from the reference conditional distribution.
If the intervention is on both the bias and the cause, $A$ is sampled from a different marginal distribution, $X$ is sampled from a different marginal distribution independently of $A$, and $Y$ is sampled from the reference conditional distribution.
If the intervention is on the effect, $A$ is sampled from the reference marginal distribution, $X$ is sampled from the reference conditional distribution, and $Y$ is sampled from a different marginal distribution independently of $A$ and $X$.
Thus, we obtain all the transfer joint distributions that arise from interventions on some of the variables (Table~\ref{table:Tranfer distribution}).

\subsection{Fairness-aware Models for Training}
\label{sec: Fairness-aware Models}

    The role of SCM mentioned earlier is to demonstrate the variables involved in the actual training of the model, as well as the causal dependencies between them. The models mentioned in this section (Fig.~\ref{fig:two_model}), which are distinct from the SCM and are used for training, are referred to as causal models and anti-causal models, respectively.
The causal model and the anti-causal model are constructed with the variables (A,X,Y). The causal model can be described as
\begin{equation}
    \bm p_{\theta_\to}(a, x, y)=\bm p_{\theta_{A}}(a)\bm p_{\theta_{X|A}}(x|a)\bm p_{\theta_{Y|A,X}}(y|a,x).
\end{equation}
Meanwhile, the anti-causal model can be described as
\begin{equation}
    \bm p_{\theta_\gets}(a,x,y)=\bm p_{\theta_{A}}(a)\bm p_{\theta_{Y|A}}(y|a)\bm p_{\theta_{X|A,Y}}(x|a,y),
\end{equation}
where the $\theta_\to$ and $\theta_\gets$ represent parameters of the two models respectively (e.g.\ $\theta_\to$ includes $\theta_{A}$, $\theta_{X|A}$, and $\theta_{Y|A,X}$).

\par
\subsection{Adaptation of Two Models}
\label{sec: adaptation}
\begin{algorithm}[!tp]
\textbf{Input:}{The initialized causal model $\bm p_{\theta_\to^{(0)}}=\bm p$, the intervened variable $\bm v$ (e.g. a,x or y), training epochs T, learning rate $\alpha$, num of samples in each epoch $\bm K$.}\\
\textbf{Goal:}Let initial parameter $\theta^{(0)}$ adapt to the parameter $\theta^{*}$ of the transfer distribution $\bm p^*$.

$p^*(a,x,y)=p(a,x,y|do(v=t))$; t represents random interference on variable v \newline
\For{t=1 to T} 
{
    \For{k=1 to K}
    {Sample $\theta_k=(a_k,x_k,y_k)\sim p^*(a,x,y)$\\
    
    }
    $\theta_K=\left \{ \theta_k \right \} _{k=0}^{K}$\\
    Calculate the loss $\cL_\text{causal}$ in Eqn.~\ref{eq:causal_loss} with $\theta_K$\\
    $\theta_t=\theta_{t-1}-\alpha\nabla_{\theta_{t-1}} \cL_\text{causal} $
}
\label{algorithm:1}
\caption{Adaptation of Causal Model}
\end{algorithm}
 This section explains the most fundamental issue of this work. Assuming the initial distribution of two models is both p, and due to certain factors, the distribution drifts to $p^*$, the training process is to make the models approach the distribution $p^*$(Algorithm~\ref{algorithm:1}).  When the training is about to start, the training term is $T:=0$. The two models will be initialized to fit the same reference distribution \textbf{p} like
    \begin{equation}
        \bm p_{\theta_\to^{(0)}}=\bm p_{\theta_\gets^{(0)}}=\bm p.
    \end{equation}
Then we get samples from the transfer distribution $p^*$. Letting these samples join the training process, the log-likelihood will gradually increase in every step of stochastic gradient descent (SGD). The distribution $\bm p_{\theta}$ adapts to $p^*$ closest until we get the minimal log-likelihood loss. Taking the causal model for example, the loss is 
\begin{align}
    \cL_\text{causal}(\causalvar)
    = &\expect[(A,X,Y) \sim \ptransfer]{-\log \pmodel(A,X,Y)}\nonumber\\ 
    = & \expect[\ptransfer]{-\log \ptrue_{\theta_A}(A)} + \expect[\ptransfer]{-\log \ptrue_{\theta_{A|X}}(A|X)} \nonumber \\
    + & \expect[\ptransfer]{-\log \ptrue_{\theta_{Y|A,X}}(Y|A,X)},
    \label{eq:causal_loss}
\end{align}
where the log-likelihood suboptimality is equal to the KL-divergence
\begin{align}
    \cL(\theta) - \cL(\theta^*) = \KL(\ptrue^* || \ptrue_\theta).
\end{align}

\section{Parameters and Analysis}
\label{sec:analysis}

\subsection{Relavant Inequality}
\label{Relavant Inequality}
Based on Average Stochastic Gradient Descent (ASGD)~\cite{bottou2012stochastic}, the previous work~\cite{2020arXiv200509136L} proves that the average parameter's $\bar \theta^{(T)} = \inv{T} \sum_{t=0}^{T-1} \theta^{(t)}$ suboptimality  is upper bounded by
\begin{align}
    \expect{\KL(\ptrue^*||\ptrue_{\bar \theta^{(T)}})}
    \leq \frac{
    c^{-1} \|\theta^{(0)} - \theta^* \|^2
     + c \smoothness^2}{2 \sqrt{T}},
    \label{eq:sgd_rate}
\end{align}
where $c$ is a small enough constant.
This inequality indicates that the upper bound of the convergence is mainly determined by the distance $\delta:= \|\theta^{(0)} - \theta^* \|^2$ between the reference distribution and the transfer distribution. Specifically, the initial distance of the causal model and the anti-causal model is respectively denoted by $\delta_\text{causal} = \|\causalvar^{(0)} - \causalvar^*\|^2$ and $\delta_\text{anticausal} = \|\antivar^{(0)} - \antivar^*\|^2$. And the two distances are the core basis of the subsequent discussion.
\begin{figure}
\centering   \includegraphics[width=\columnwidth]{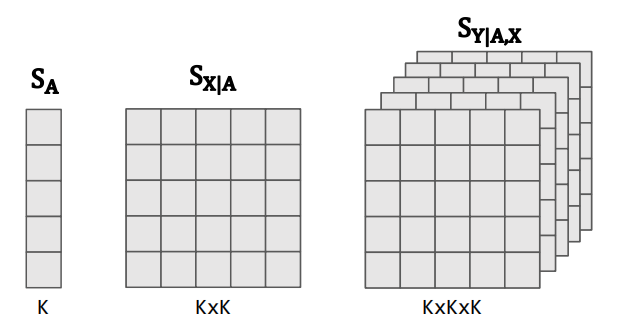}
\caption{Parameters of the causal model in SGD. In which the length K of each dimension represents that the variables A, X, and Y can be divided into K categories.}
\label{fig: parameters}
\end{figure}

\subsection{Parameters in Trainable Models}
\label{Sec: Parameters}
We assume that the bias $A$, the cause $X$, and the effect $Y$ in the two models are multiclass variables with $K$ classes. The natural parameter $\boldsymbol{\theta}\in\mathbf{R}^K$ is generated from the distribution $p$ by the inverse function of the softmax function
\begin{align}
    \evp_z = \frac{e^{\evs_z}}{\sum_{z'} e^{\evs_{z'}}}.
\end{align}
We set $\vs$ as the trainable parameter in SGD. Taking causal model for example, the model has parameters $\vs_A := (\evs_a)_{a=1\dots K}$, $\vs_{X|A} := (\evs_{x|a})_{a,x=1\dots K}$
and $\vs_{Y|A,X} := (\evs_{y|a,x})_{a,x,y=1\dots K}$.  The parameters of causal model can be represented as $\causalvar = (\vs_A, \vs_{X|A}, \vs_{Y|A,X})$, while that's $\antivar = (\vs_A, \vs_{Y|A}, \vs_{X|A,Y})$ in anti-causal model. We describe the relationship between the parameter shapes in Fig.~\ref{fig: parameters}. Using the parameter $\vs$, the loss (\ref{eq:causal_loss}) can be written as 
\begin{align}
    \cL_\text{causal}(\causalvar) 
    & = \expect[(A,X,Y)\sim\ptransfer]{-\log \pmodel(A, X, Y)}\nonumber\\
    & = \mathbf{E}_{\ptransfer}[ - \evs_A + \log\sum_{a} e^{s_{a}} 
    -  \evs_{X|A} + \log\sum_x e^{s_{x|A}}\nonumber\\ 
    & - \evs_{Y|A,X} + \log\sum_y e^{s_{y|A,X}}].
\end{align}
After deriving the new form of the loss, we can use inequality (\ref{eq:sgd_rate}) to further compare the adaptation speeds of the two models. In other words, we only need to compare the initial distances $\delta_\text{causal}$ and $\delta_\text{anticausal}$ as described in Section~\ref{Relavant Inequality}.
\subsection{Adaptation Speeds of Two Models}
Based on the introduction in Section \ref{Relavant Inequality}, the adaptation speed of two models depends on the distance between their initial distribution and the distribution after the change, which can be represented by the parameter "s" in Section \ref{Sec: Parameters}. Furthermore, comparing the convergence speed of two models can be equivalently expressed as comparing the magnitudes of $\delta_\text{causal} = \|\causalvar^{(0)} - \causalvar^*\|^2$ and $\delta_\text{anticausal} = \|\antivar^{(0)} - \antivar^*\|^2$, the model with a smaller initial distance $(\delta)$ has a faster convergence speed.

\textbf{Domain Shift by bias A,} $\vs_A \assign \vs^*_A$. In this scenario, both the causal and anti-causal models modify only the same sensitive marginal $\vs_A$. The initial distance between the two models can be expressed as
\begin{align}
\delta_\text{causal} = \delta_\text{anticausal} = \|\vs_A - \vs^*_A\|^2,
\end{align}
which implies that the two models converge simultaneously.

\textbf{Domain Shift by cause X,} $\forall a, \vs_{X|a} \assign \vs^*_X$. The conditional $\vs_{Y|A,X}$ remains unchanged while $\vs_{Y|A}$ and $\vs_{X|A,Y}$ of anti-causl model are modified.The initial distance can be written as 
\begin{align}
    \delta_\text{causal} &= \sum_a\|\vs_{X|a} - \vs^*_X\|^2, \\
    \delta_\text{anticausal}
    &= \sum_a\|\vs_{Y|a} - \vs^*_{Y|a}\|^2  + \sum_a\sum_y \|\vs_{X|a,y} - \vs^*_{X|a,y}\|^2.
\end{align}
We can compare the initial distances of the two models based on the aforementioned distance and arrive at the following proposition.
\begin{proposition}
\label{pro:cause}
When the intervention is on the cause,
\begin{equation}
    \delta_\text{anticausal} \geq K \delta_\text{causal} \; ,
    \label{equ:compare1}
\end{equation}
where the specific proof process will be explained in Appendix~\ref{sec:proof for p1}.
\end{proposition}

\textbf{Domain Shift by both bias A and cause X,} $ \vs_{A} \assign \vs^*_A$ and  \\$ \forall a,\vs_{X|a} \assign \vs^*_X$. 
Compared to the case of intervening on cause X, the intervention on bias A introduces an additional equal distance to the initial distance of the two models. Hence, we can obtain a similar result as before and derive the following inequality:
\begin{equation}
    \delta_\text{anticausal} \geq \delta_\text{causal} \;,
    \label{eql:bias_and_cause}
\end{equation}
which will be simply explained in Appendix~\ref{sec:proof for ax}.
It can be observed that in all three scenarios, the initial distance of the causal model is consistently smaller than that of the anti-causal model, indicating that the causal model adapts to the domain more quickly. However, if the domain shift is induced by interventions on the effect variable (Y), interestingly, different conclusions can be drawn.

\textbf{Domain Shift by effect Y,} $ \forall a,x ,\vs_{Y|a,x} \assign \vs^*_Y $. The marginal $\vs_{A}$ and the conditional $\vs_{X|a}$ remain unchanged under this intervention. On the other hand, the conditional  $\vs_{Y|A}$ and $\vs_{Y|A,X}$ in the anti-causal model change with respect to effect Y. The initial distances for the two models can be expressed as:
\begin{align}
        &\delta_\text{causal} = \|\vs_{Y} - \vs^*_{Y|A,X}\|^2,\\
        &\delta_\text{anticausal}
    = \sum_a \|\vs_{Y|a} - \vs^*_{Y}\|^2
     + \sum_a\sum_y \|\vs_{X|a,y} - \vs^*_{X|a,y}\|^2,
\end{align}
where the distance above does not maintain a constant relationship compared to the other three situations.

\begin{proposition}
\label{pro:effect}
When the intervention is on the effect, 
there will be two situations. If the following inequality: \begin{equation}
     \|\vs^*_Y - \vc\|^2 < R^2
     \label{fomula:pro2}
\end{equation} is satisfied, $    \delta_\text{anticausal} \geq  \delta_\text{causal} \; $ (proved in Appendix~\ref{sec:proof for p2}).
Let us illustrate this inequality,
where $R^2$ (see Appendix for specific formula) is a constant related with $\vs_{X}$, $\vs_{Y|A}$, $\vs_{X|A,Y}$ and $\vs_{Y|A,X}$
, and $\vc = \frac{\left(\sum_x \vs_{Y|A,x}\right) - \vs_{Y|A} }{K-1}$.
\end{proposition}
The inequality (\ref{fomula:pro2}) implies that the causal model has a comparative advantage only within a certain range, where the modified marginal $\vs_{Y}$ is sufficiently close to $\vc$. However, if the distance goes beyond this range, the anti-causal model will converge faster.

\section{Experiments}
\label{sec:exp}

\begin{table*}[!t]
  \centering
  \begin{tabular}{ | c | c  c  c | }
    \hline
     Intervention & \makecell[c]{Scatter Plot\\T=500} & \makecell[c]{Scatter Plot\\T=1500} & Convergence Curve
     \\ 
         Bias&
        \begin{minipage}[b]{0.555\columnwidth}
		\centering
		\raisebox{-.5\height}{\includegraphics[width=\linewidth]{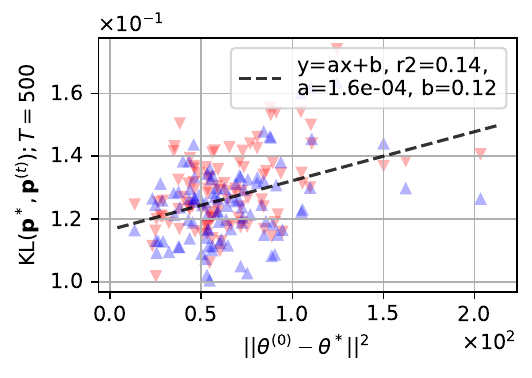}}
	\end{minipage}
        &
        \begin{minipage}[b]{0.555\columnwidth}
        \centering
		\raisebox{-.5\height}{\includegraphics[width=\linewidth]{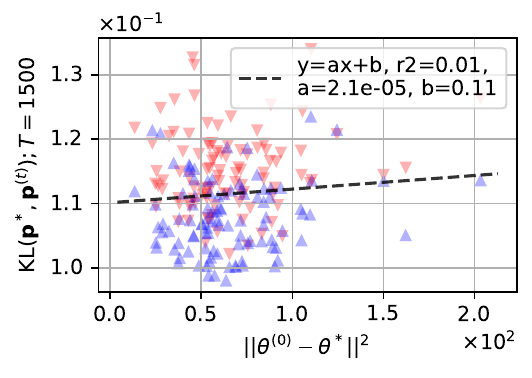}}
        \end{minipage}
        &
	\begin{minipage}[b]{0.555\columnwidth}
		\centering
		\raisebox{-.5\height}{\includegraphics[width=\linewidth]{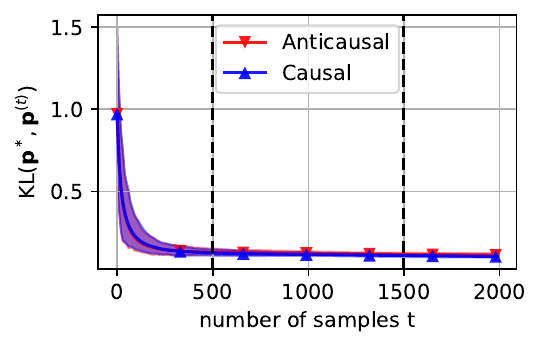}}
        \end{minipage}
    \\ 
        Cause&
        \begin{minipage}[b]{0.555\columnwidth}
		\centering
		\raisebox{-.5\height}{\includegraphics[width=\linewidth]{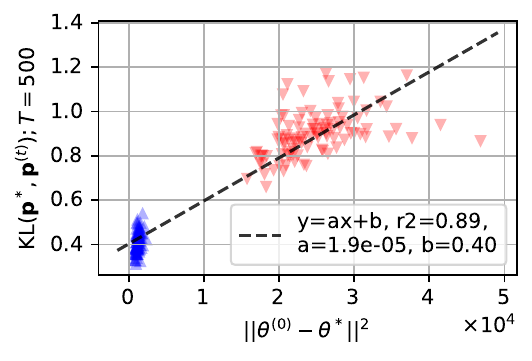}}
	\end{minipage}
        &
        \begin{minipage}[b]{0.555\columnwidth}
        \centering
		\raisebox{-.5\height}{\includegraphics[width=\linewidth]{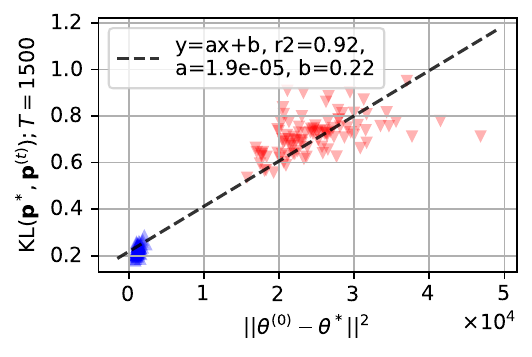}}
        \end{minipage}
        &
	\begin{minipage}[b]{0.555\columnwidth}
		\centering
		\raisebox{-.5\height}{\includegraphics[width=\linewidth]{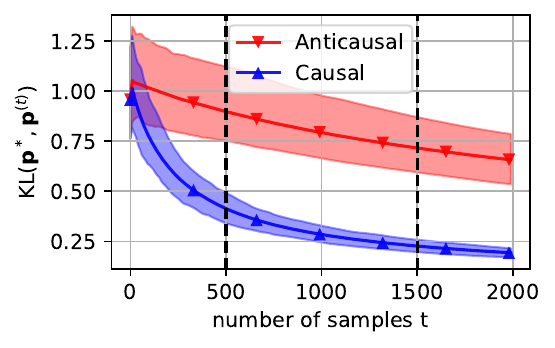}}
        \end{minipage}
        \\ 
        \makecell[c]{Bias \\ and \\Cause} &
        \begin{minipage}[b]{0.555\columnwidth}
		\centering
		\raisebox{-.5\height}{\includegraphics[width=\linewidth]{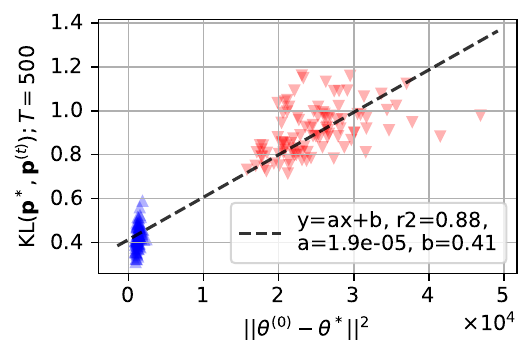}}
	\end{minipage}
        &
        \begin{minipage}[b]{0.555\columnwidth}
        \centering
		\raisebox{-.5\height}{\includegraphics[width=\linewidth]{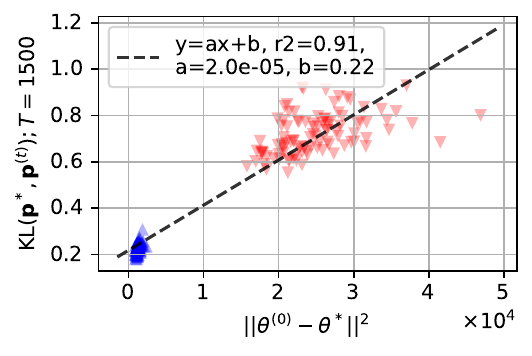}}
        \end{minipage}
        &
	\begin{minipage}[b]{0.555\columnwidth}
		\centering
		\raisebox{-.5\height}{\includegraphics[width=\linewidth]{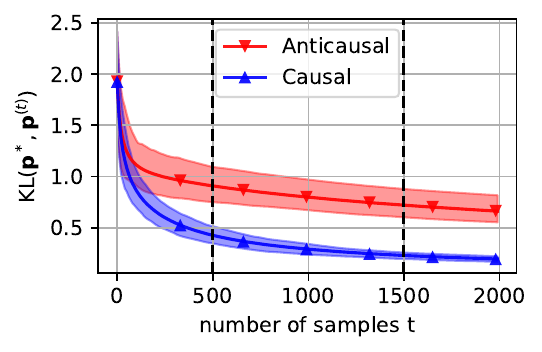}}
        \end{minipage}
        \\ 
        Effect&
        \begin{minipage}[b]{0.555\columnwidth}
		\centering
		\raisebox{-.5\height}{\includegraphics[width=\linewidth]{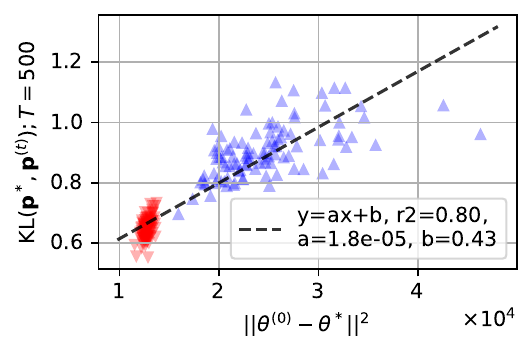}}
	\end{minipage}
        &
        \begin{minipage}[b]{0.555\columnwidth}
        \centering
		\raisebox{-.5\height}{\includegraphics[width=\linewidth]{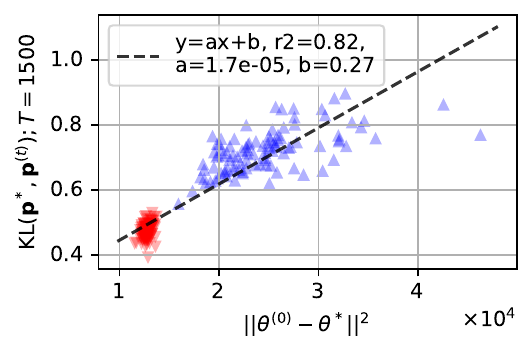}}
        \end{minipage}
        &
	\begin{minipage}[b]{0.555\columnwidth}
		\centering
		\raisebox{-.5\height}{\includegraphics[width=\linewidth]{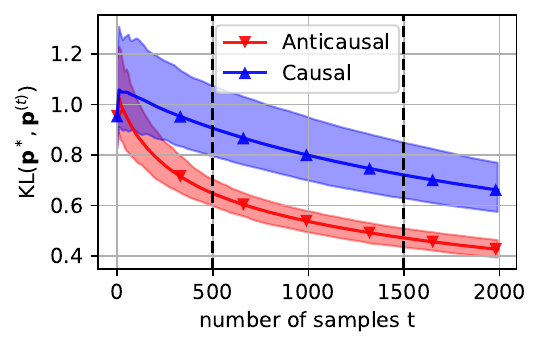}}
        \end{minipage}
        \\ \hline
  \end{tabular}
   \caption{\textbf{Results on synthetic data.} The scatter plots in the first column demonstrate the positive correlation between the KL-divergence after one-quarter of the training steps, while the second column shows the correlation after three-quarters of the training steps. Each point on the scatter plots represents a pair $(p^{(0)},p^*)$ in the causal model (blue) or the anti-causal model (red). The parameters a, b, and $r^2$ in a scatter plot represent the slope, intercept, and the coefficient of determination, respectively, in the least squares method. The coefficient of determination measures the linear correlation between the independent and dependent variables. The curve in the third column shows the relative speeds of the two models. The shaded area indicates the 5th and 95th percentiles of the KL-divergence. The value of $K_{synthetic}$ is set to 20 in this experiment. }
    \label{table:synthetic result}
\end{table*}
\begin{table*}[!t]
  \centering
  \begin{tabular}{ | c | c  c  c | }
    \hline
    Intervention & \makecell[c]{Scatter Plot\\T=500} & \makecell[c]{Scatter Plot\\T=1500} & Convergence Curve
     \\ 
         Bias&
        \begin{minipage}[b]{0.555\columnwidth}
		\centering
		\raisebox{-.5\height}{\includegraphics[width=\linewidth]{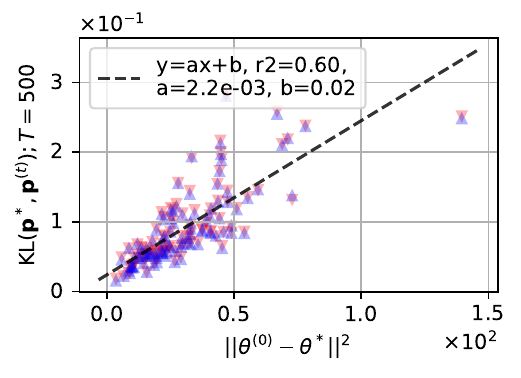}}
	\end{minipage}
        &
        \begin{minipage}[b]{0.555\columnwidth}
        \centering
		\raisebox{-.5\height}{\includegraphics[width=\linewidth]{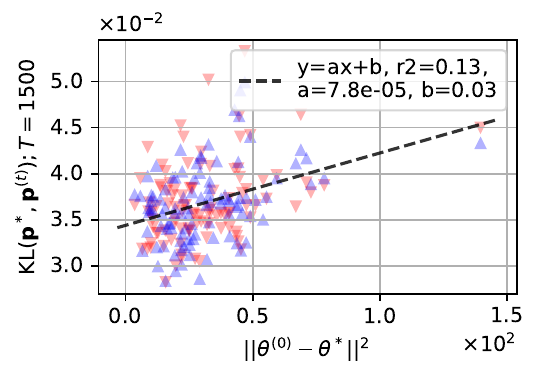}}
        \end{minipage}
        &
	\begin{minipage}[b]{0.555\columnwidth}
		\centering
		\raisebox{-.5\height}{\includegraphics[width=\linewidth]{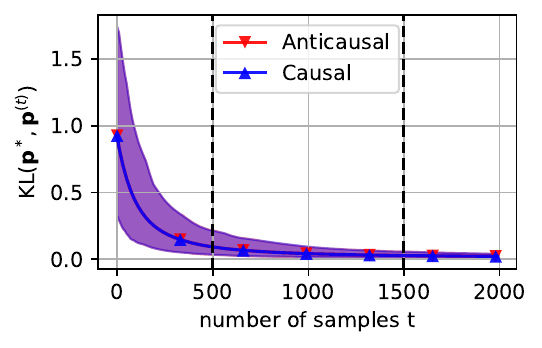}}
        \end{minipage}
    \\ 
        Cause&
        \begin{minipage}[b]{0.555\columnwidth}
		\centering
		\raisebox{-.5\height}{\includegraphics[width=\linewidth]{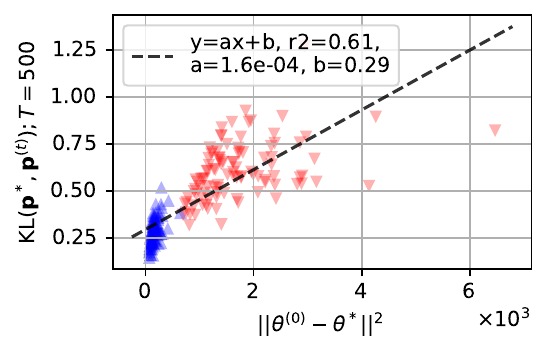}}
	\end{minipage}
        &
        \begin{minipage}[b]{0.555\columnwidth}
        \centering
		\raisebox{-.5\height}{\includegraphics[width=\linewidth]{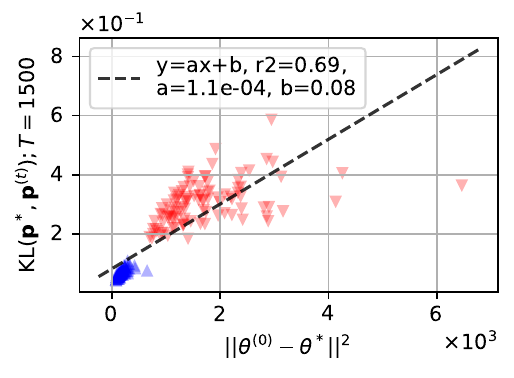}}
        \end{minipage}
        &
	\begin{minipage}[b]{0.555\columnwidth}
		\centering
		\raisebox{-.5\height}{\includegraphics[width=\linewidth]{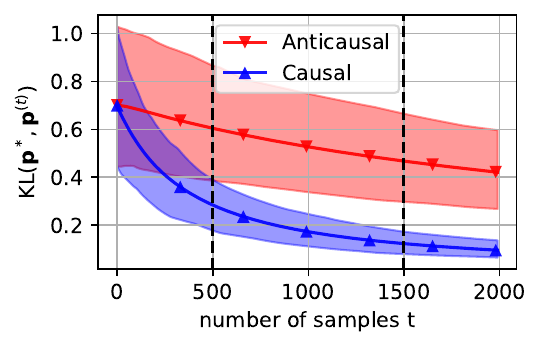}}
        \end{minipage}
        \\ 
        \makecell[c]{Bias \\ and \\Cause} &
        \begin{minipage}[b]{0.555\columnwidth}
		\centering
		\raisebox{-.5\height}{\includegraphics[width=\linewidth]{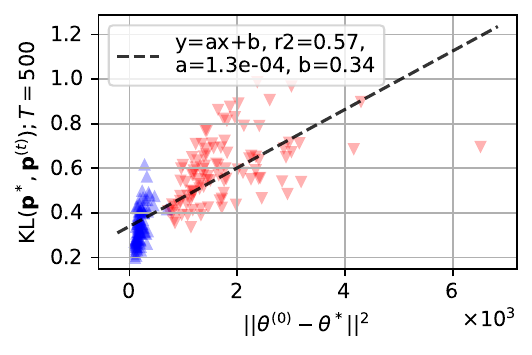}}
	\end{minipage}
        &
        \begin{minipage}[b]{0.555\columnwidth}
        \centering
		\raisebox{-.5\height}{\includegraphics[width=\linewidth]{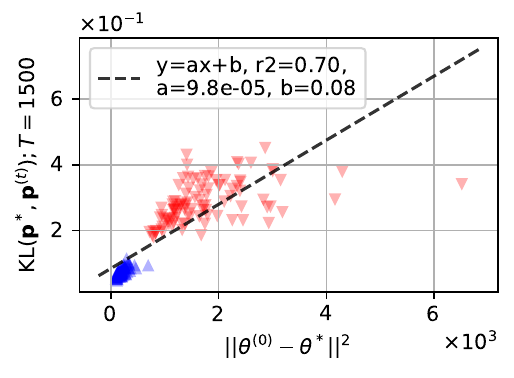}}
        \end{minipage}
        &
	\begin{minipage}[b]{0.555\columnwidth}
		\centering
		\raisebox{-.5\height}{\includegraphics[width=\linewidth]{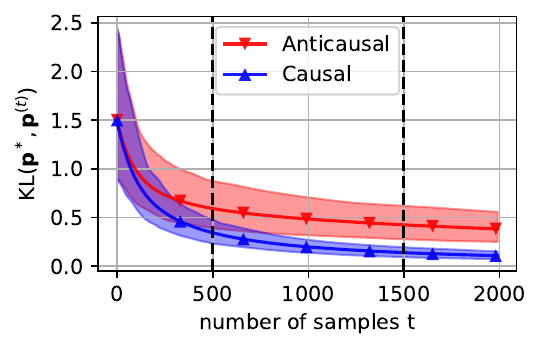}}
        \end{minipage}
        \\ 
        Effect&
        \begin{minipage}[b]{0.555\columnwidth}
		\centering
		\raisebox{-.5\height}{\includegraphics[width=\linewidth]{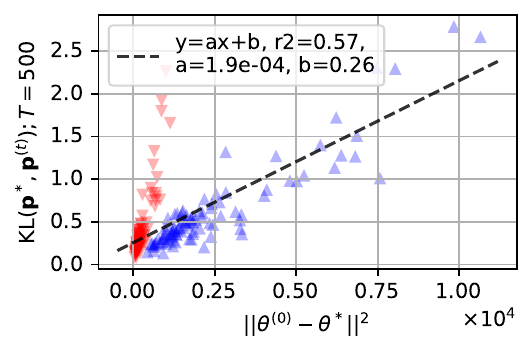}}
	\end{minipage}
        &
        \begin{minipage}[b]{0.555\columnwidth}
        \centering
		\raisebox{-.5\height}{\includegraphics[width=\linewidth]{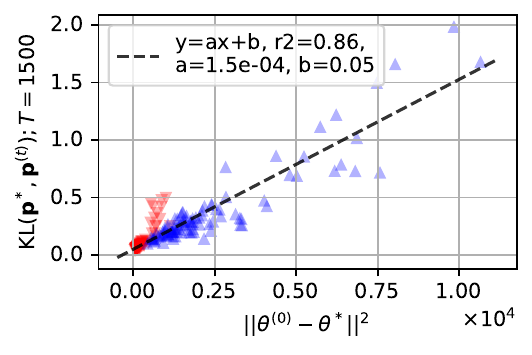}}
        \end{minipage}
        &
	\begin{minipage}[b]{0.555\columnwidth}
		\centering
		\raisebox{-.5\height}{\includegraphics[width=\linewidth]{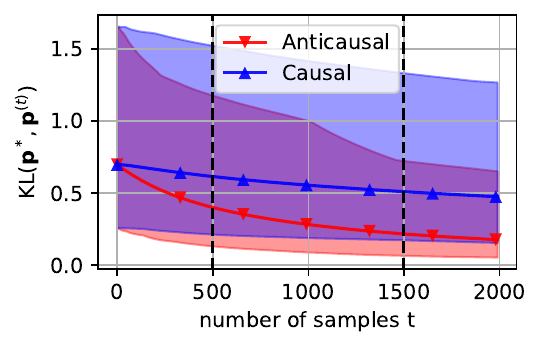}}
        \end{minipage}
        \\ \hline
  \end{tabular}
  \caption{\textbf{Results on Color-MNIST data.} The relative positional relationship between the two types of points in the scatter plots remains unchanged. Here we set $K_{MNIST}=10$. The convergence speed of the two models is consistent with previous experiments on synthetic data.}
    \label{table:mnist result}
\end{table*}
\begin{table*}[!t]
  \centering
  \begin{tabular}{ | c | c  c  c | }
    \hline
     Intervention & \makecell[c]{Scatter Plot\\T=50} & \makecell[c]{Scatter Plot\\T=150} & Convergence Curve
     \\ 
         Bias&
        \begin{minipage}[b]{0.555\columnwidth}
		\centering
		\raisebox{-.5\height}{\includegraphics[width=\linewidth]{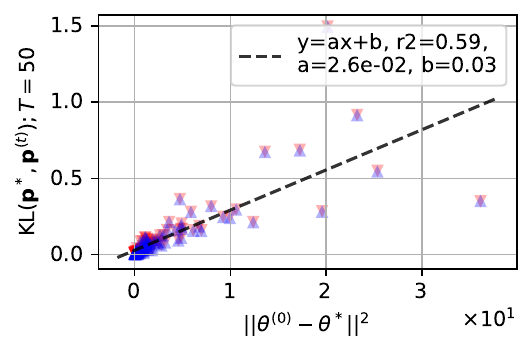}}
	\end{minipage}
        &
        \begin{minipage}[b]{0.555\columnwidth}
        \centering
		\raisebox{-.5\height}{\includegraphics[width=\linewidth]{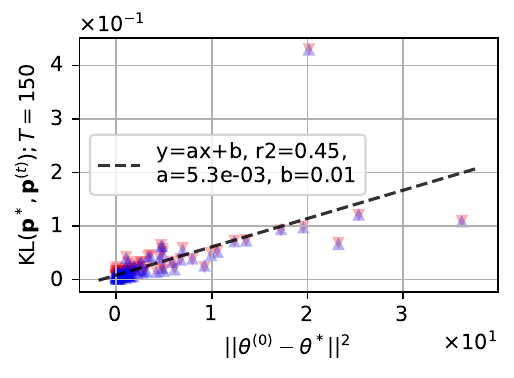}}
        \end{minipage}
        &
	\begin{minipage}[b]{0.555\columnwidth}
		\centering
		\raisebox{-.5\height}{\includegraphics[width=\linewidth]{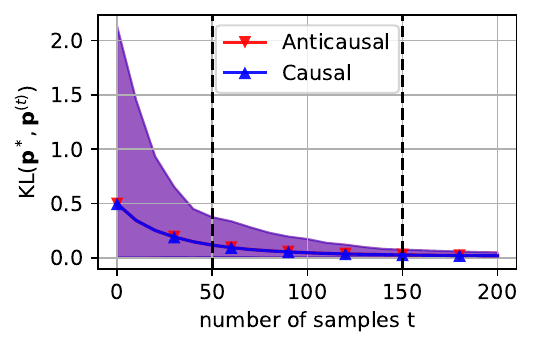}}
        \end{minipage}
    \\ 
        Cause&
        \begin{minipage}[b]{0.555\columnwidth}
		\centering
		\raisebox{-.5\height}{\includegraphics[width=\linewidth]{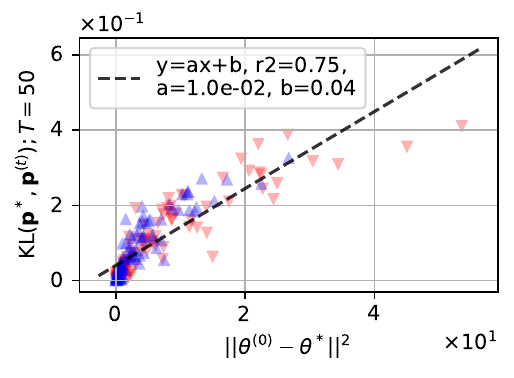}}
	\end{minipage}
        &
        \begin{minipage}[b]{0.555\columnwidth}
        \centering
		\raisebox{-.5\height}{\includegraphics[width=\linewidth]{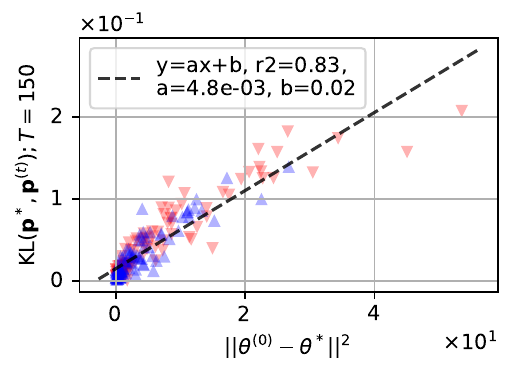}}
        \end{minipage}
        &
	\begin{minipage}[b]{0.555\columnwidth}
		\centering
		\raisebox{-.5\height}{\includegraphics[width=\linewidth]{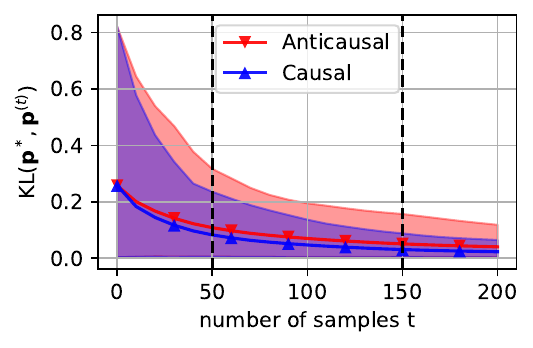}}
        \end{minipage}
        \\ 
        \makecell[c]{Bias \\ and \\Cause} &
        \begin{minipage}[b]{0.555\columnwidth}
		\centering
		\raisebox{-.5\height}{\includegraphics[width=\linewidth]{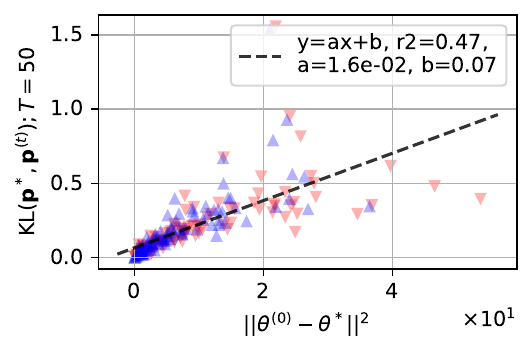}}
	\end{minipage}
        &
        \begin{minipage}[b]{0.555\columnwidth}
        \centering
		\raisebox{-.5\height}{\includegraphics[width=\linewidth]{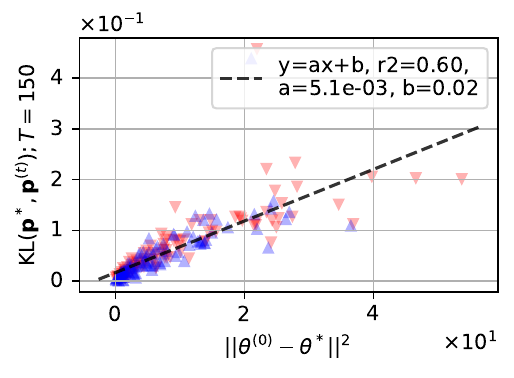}}
        \end{minipage}
        &
	\begin{minipage}[b]{0.555\columnwidth}
		\centering
		\raisebox{-.5\height}{\includegraphics[width=\linewidth]{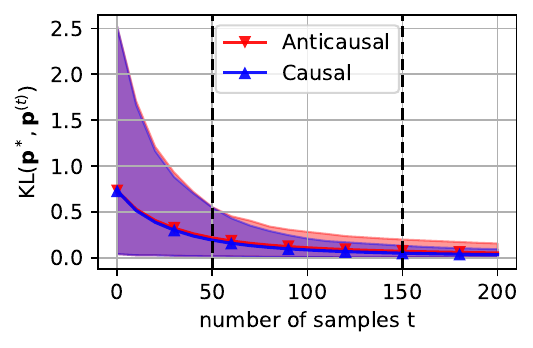}}
        \end{minipage}
        \\ 
        Effect&
        \begin{minipage}[b]{0.555\columnwidth}
		\centering
		\raisebox{-.5\height}{\includegraphics[width=\linewidth]{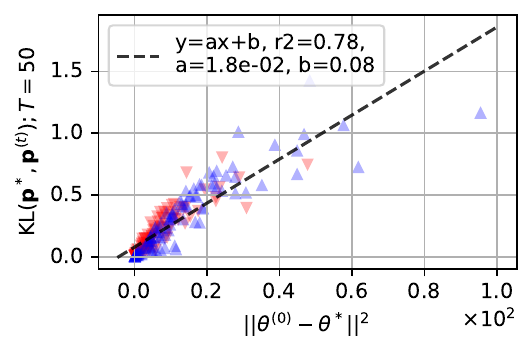}}
	\end{minipage}
        &
        \begin{minipage}[b]{0.555\columnwidth}
        \centering
		\raisebox{-.5\height}{\includegraphics[width=\linewidth]{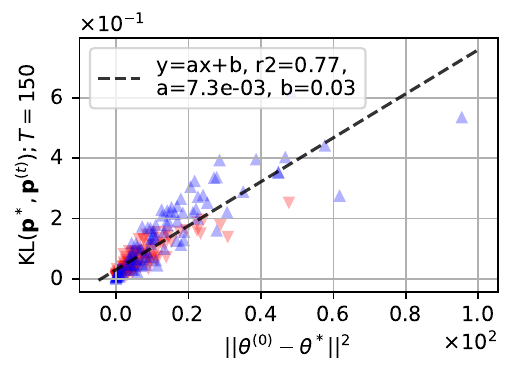}}
        \end{minipage}
        &
	\begin{minipage}[b]{0.555\columnwidth}
		\centering
		\raisebox{-.5\height}{\includegraphics[width=\linewidth]{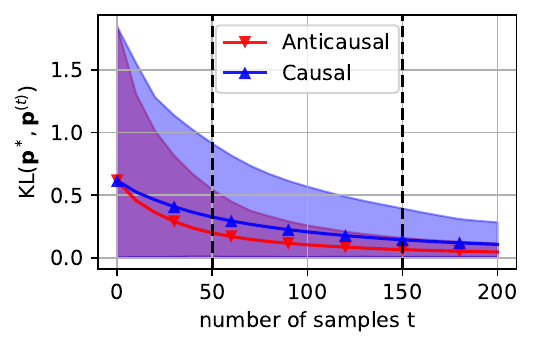}}
        \end{minipage}
        \\ \hline
  \end{tabular}
  \caption{\textbf{Results on the Adult data.} As an experiment using binary variables, we observed that both the causal and anti-causal models converge faster, similar to the results in the previous experiments. Additionally, the two adaptation curves are closer when $K_{Adult}=2$. Furthermore, there is less separation between the points of the two models in the scatter plots.}
    \label{table:adult result}
\end{table*}
\subsection{Data and Settings}
\label{c1}

\textbf{Synthetic Data.} The first we need is to get the distributions $ \bm{p= p_{\theta^{(0)}}}$ which is called \textit{prior}. Specifically, we get $\bm{p}_A$, $\bm{p}_{X|A}$ and $\bm{p}_{Y|A,X}$ from the Dirichlet distribution. The three distributions can be represented as: 
\begin{align*}
    \bm{p}_A &\sim Dirichlet (\bm{1}_K), \\
    \forall a, \bm{p}_{X|a} &\sim Dirichlet (\bm{1}_K), \\
    \forall a,x, \bm{p}_{Y|a,x} &\sim Dirichlet (\bm{1}_K),
\end{align*}
where $\bm{1}_K$ is the all-one vector of K-dimension. We can say that by using the aforementioned distributions, we ensure that these three distributions are mutually independent. Now we have thus obtained an initial joint distribution.
\begin{figure}
    \centering   \includegraphics[width=\columnwidth]{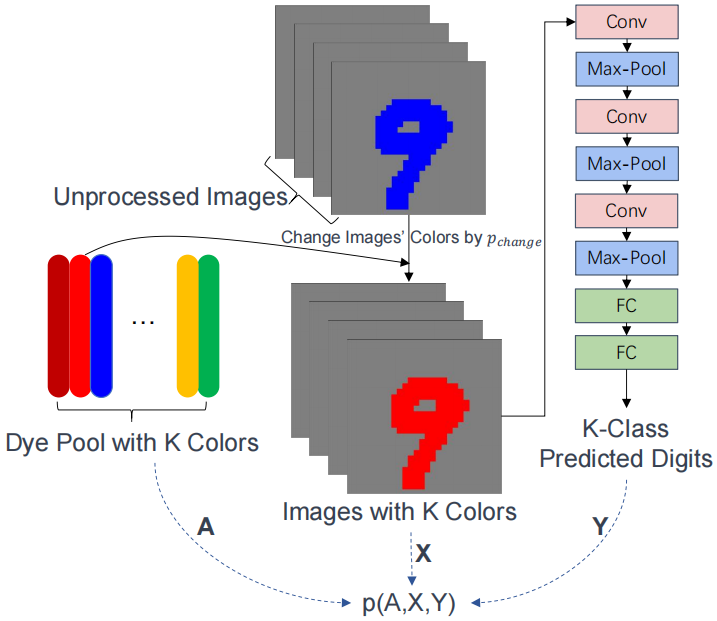}
    \caption{Initialize p(A,X,Y) with Colored-MNIST. The color pool consists of K RGB colors, and it can probabilistically alter the colors of the numbers in the image. The image is also segmented into K categories based on its color, and K prediction results are generated by CNN and linear layer.
    }
    \label{fig:color-mnist}
\end{figure}

Besides employing synthetic data, we integrated two actual datasets to establish the initial distribution $\bm p$. This involved determining the frequency of each category for each variable. In the subsequent sections, we will elucidate the identifiable variables A, X, and Y present in each respective dataset. It is important to note that we did not directly train the causal model and anti-causal model using the adult and colored-MNIST datasets, nor did we compare their convergence speeds in this manner. Instead, all experiments in this article were conducted employing the two models described in Section~\ref{sec: Fairness-aware Models}, with the parameters mentioned in Section~\ref{Sec: Parameters}, and the convergence speeds were compared using the algorithm outlined in Section~\ref{sec: adaptation}. To differentiate it from the strictly Dirichlet distribution in synthetic data, we aimed to acquire the initial distribution $p(A,X,Y)$ from real datasets to showcase additional results.

\par
\textbf{Colored-MNIST.} We use MNIST~\cite{726791} to construct the initial distribution $\bm p$. The sensitive variable (A) is RGB color values, which consist of K kinds of colors represented by a three-dimensional vector. We obtain the initial $\bm{p}_{color}$ from the Dirichlet distribution. The cause (X) is the images' features of colored-MNIST, which have different colors and can be classified into K categories based on color. For each image, we randomly select a color from $\bm{p}_{color}$ to stain it. To avoid having the same distribution $\bm{p}_{image}$ as $\bm{p}_{color}$, we set the probability of changing the image's color as $\bm p_{change}$(e.g., $\bm p_{change}=0.5$). Then we need a model to input image features (X) and get predicted numbers (Y). Since we only need the initial distribution of Y obtained from X, the model used to obtain the predicted numbers is not crucial in this analysis. We process the images with a convolutional neural network (CNN) to predict the numbers, which represent the effect (Y) of the SCM. Finally, we obtain the initial distribution $\bm p(A,X,Y)$ by calculating the frequency of all categories (Fig.\ref{fig:color-mnist}).

\textbf{Adult.} The results from the MNIST dataset and synthetic data demonstrate the adaptation speed of multi-categorical variables. We use the Adult dataset ~\cite{1997Scaling} to demonstrate the adaptation speed of binary variables, where the bias (A) is gender, including male and female. The initial distribution $\bm{p}_{gender}$ is also obtained from the Dirichlet distribution. The cause (X) is the features of people in the dataset, which are divided into two categories based on gender. And for each individual, we select a gender from $\bm{p}_{gender}$. Similarly to the Colored-MNIST dataset, we change the initial gender of each feature to the selected gender with the probability $\bm p_{change}$. Using the modified features, we input them into a 2-layer MLP to obtain binary predicted labels. The binary labels are recognized as the effect (Y) in the SCM.

\subsection{Results}

This section demonstrates the adaptation speed of the causal and anti-causal models using the data from Section \ref{c1}, as well as the positive relationship between the KL divergence and the initial parameter distance. And similar results were obtained in both multi-class and binary classification datasets.

Utilizing $K_{\text{synthetic}}=20$, $K_{\text{MNIST}}=10$, and $K_{\text{Adult}}=2$, the outcomes for the three datasets are presented in Table~\ref{table:synthetic result}, Table~\ref{table:mnist result}, and Table~\ref{table:adult result}, respectively, to showcase the adaptation speed of the causal and anti-causal models. Additionally, scatter plots are employed to illustrate the positive correlation between the KL divergence and the initial parameter distance throughout the training process. This positive correlation holds significant importance as it validates Inequality~\ref{eq:sgd_rate}, which constitutes the foundation for deriving the relationship between the model's adaptation speed. The results after one-quarter and three-quarters of the training steps are presented, while the curves depict the adaptation speed.
\par
\textbf{Domain Shift by bias A.}
When intervening on bias A, both the causal and anti-causal models undergo the same change concerning the bias, while holding other variables constant. Therefore, both models exhibit overlapping points on the scatter plots, resulting in the initial distance $\delta_\text{causal} =\delta_\text{anticausal}$ and coinciding curves in the third column. As a result, the convergence speed of both models is the same. This is in line with the observation that interventions on bias do not change the causal structure of the model, and thus do not provide any new information that can be used to distinguish between the two models.

\textbf{Domain Shift by cause X.} In the case of an intervention on the cause X, the causal model has an advantage over the anti-causal model, as the points of the causal model cluster towards the bottom left as compared to the points of the anti-causal model. This relative positioning reflects the formula $\delta_\text{causal} < \delta_\text{anticausal}$, which is derived in our previous work. This advantage of the causal model is most pronounced in four intervention scenarios, where the causal model exhibits a significant advantage in terms of the curves.

\textbf{Domain Shift by both bias A and cause X.} When intervening on both the bias and the cause, the relative positions of points in the causal and anti-causal models maintain their relative positions, with the anti-causal model being on the upper right of the causal model as a whole. However, the difference in adaptation speed between the two models shrinks, as the two curves have a short overlap in the early training steps. This result is somewhat counterintuitive, as one might expect that intervening on both the bias and the cause would provide more information that could be used to distinguish between the two models. However, our results suggest that this is not always the case, and that the advantage of the causal model may depend on the specific structure of the model.

\textbf{Domain Shift by effect Y.} When intervening on the effect Y, the relative positions of points in the causal and anti-causal models are reversed, with the points of the anti-causal model being concentrated in the lower-left corner. This advantage of the anti-causal model corresponds to the proposition that the anti-causal model is better suited to handle interventions on effect variables. Although the anti-causal model may not always have this advantage, it typically retains it in most cases. The curves also reflect this result, as the anti-causal model consistently converges faster when using the three datasets. Our results suggest that interventions on effect variables provide valuable information that can be used to distinguish between the causal and anti-causal models.

In addition to the analysis of multiple categorical variables, we also conducted a study on dichotomous variables using the Adult dataset. The results, as presented in Table~\ref{table:adult result}, indicate that both models exhibit accelerated convergence rates due to the reduced number of parameters in the training process. Except the bias intervention, the points of both models tend to converge and become less separated in the scatter plots in other scenarios. Interestingly, the relationship between the speed of the causal model and the anti-causal model remains consistent with the results from the previous two experiments. However, the two curves are significantly closer together in this study, suggesting that the number of categories of a variable may only affect the absolute adaptation, but not the relative speed of the causal model and anti-causal model.
Based on these observations, we can conclude that dichotomous variables also have a significant impact on the performance of causal models. By reducing the complexity of the models, the convergence rate can be accelerated, and the overall accuracy of the models can be improved. However, the relationship between the causal model and the anti-causal model remains unchanged, indicating that these two models are equally capable of capturing the underlying causal relationships in the data, irrespective of the type of variable used.

Returning to the example introduced in the abstract of this article, suppose that Japanese is a language developed from Chinese, in which Chinese is a causal factor in the development of Japanese. In the presence of the sensitive variable of polysemy, training a Japanese corpus using Chinese data ( training in the causal direction of the variable) may not necessarily result in faster adaptation to domain shift compared to training a Chinese corpus using Japanese data (  training in the anti-causal direction of the variable). This depends primarily on how domain shift occurs, i.e., which variable(s) undergo intervention. 

\section{Conclusion}
\label{sec:conclusion}

This paper aims to explore spurious relationships in structural causal models (SCMs) that arise due to sensitive factors. We investigate the adaptation speed of both causal and anti-causal models in the presence of bias and build upon a theory that explains the relationship between the initial distance of parameters and the adaptation speed.
Furthermore, it's a challenge to extend our analysis to models with more complex structural differences, such as those with varying numbers of variables and edges. To analyze the adaptation speed in such cases, an indicator to measure the difference between the models is necessary. However, finding an appropriate indicator can be challenging to analyze in the future.

\begin{acks}
This work is supported by NSFC program (No. 62272338).
\end{acks}

\appendix
\appendix
\section{PARAMETER}The score $\vs\in \mathbf{R} ^K$ have one additional degree of freedom compared to the probability~\cite{2020arXiv200509136L}. If we add or subtract any multiple of $\mathbf{l}=(1,1,...,1)$ to $\vs$, the softargmax will not change. That means all the scores $\{ \vs + \lambda \mathbf{l} | \forall \lambda \in \real \}$ are equivalent. Thus we can subtract the score's mean and give the scores a good property: $\sum_z \evs_z = 0$.
\section{DISTANCE ANALYSIS}
Due to space limitations, we are unable to provide complete proof, such as Lemma~\ref{lem:categorical_reverse} and the full details of the distance comparison.
\subsection{Related Variables}
It's useful to find the relation of $\bm s_{y|a,x}$ and $\bm s_{x|a,y}$, because $\delta_\text{causal}$ and $\delta_\text{anticausal}$ are represented by $\bm s$.
\begin{definition}[Part-average conditional score vectors]
\label{def:average_conditional_logit}
For any $a,x$ or $y$, define
\begin{align}
     \theta_1(a,y):= \frac{1}{K} \sum_x \evs_{y|a,x}, \quad
     \theta_2(a,x):= \frac{1}{K} \sum_y \evs_{x|a,y} \; .
\end{align}
\end{definition}

\begin{definition}[Conditional log-partition function]
    \begin{align}
        \logpartition_1(a,x) = \log \sum_y e^{\evs_{y|a,x}}\quad
        \logpartition_2(a) = \log \sum_x e^{\evs_{x|a}}
    \end{align}
    \label{Conditional log-partition function}
\end{definition}
\begin{lemma}[Anticausal conditional score]
\label{lem:categorical_reverse}
$\evs_{y|a,x}$,$\evs_{x|a,y}$, $\evs_{x|a}$ are all conditional scores in two models. Then we find the relation between $\evs_{y|a,x}$ and $\evs_{x|a,y}$.
\begin{equation}
    \label{eq:bayes_scores}
    \boxed{\evs_{x|a,y} = \evs_{y|a,x} + \evs_{x|a}-A_1(a,x)-\theta_1(a,y)+ \alpha(a),
    }
\end{equation}
where $\alpin = \frac{1}{K} \sum_x \logpartition_1(x)$.
\end{lemma}

\subsection{Proof for Proposition~\ref{pro:cause}}
\label{sec:proof for p1}
Given that $\evs^*_{y|a,x} = \evs_{y|a,x}, \theta^*_1=\theta_1, \logpartition_1^*=\logpartition_1$ and $\alpin^*=\alpin$,  Lemma~\ref{lem:categorical_reverse} tells us that D-value between the anti-causal conditional $\vs^*_{X|A,Y}$ and the causal conditional $\vs^*_{X|A}$ remains unchanged.
\begin{align*}
    \evs_{x|a,y}^{*}-\evs_{x|a}^{*}&=\evs_{y|a,x}-A_{1}(a,x)+\alpha_{1}(a)-\theta_1(a,y)\\
    &=\evs_{x|a,y}-\evs_{x|a}\\
    \implies
    \evs_{x|a,y} - \evs^*_{x|a,y}  
    &= \evs_{x|a}  -\evs^*_{x|a}   \; 
\end{align*}
The distance between models before and after intervention are
\begin{align*}
\delta_{causal}
&=\sum_{a}\left\|\evs_{X|a}-\evs_{X}^{*}\right\|^{2}
=\sum_{a,x}(\evs_{x|a}-\evs_{x}^{*})^{2}\\
   \delta_{anticausal}&=\sum_{a}\left\|\evs_{Y|a}-\evs_{Y|a}^{*}\right\|^{2}+\sum_{a,y}\left\|\evs_{X|a,y}-\evs_{X|y}^{*}\right\|^{2}\\ 
&=\sum_{a}\left\|\evs_{Y|a}-\evs_{Y|a}^{*}\right\|^{2}+\sum_{a,x,y}(\evs_{x|a}-\evs_{x|a}^{*})^{2} \\
&\ge 0+K\sum_{a,x}(\evs_{x|a}-\evs_{x}^{*})^{2}= K\delta_{causal}.
\end{align*}

\subsection{Proof for Equation~\ref{eql:bias_and_cause}}
\label{sec:proof for ax}
\begin{align*}
    \delta_{causal}
&=\left\|\evs_{A}-\evs_{A}^{*}\right\|^{2}+\sum_{a}\left\|\evs_{X|a}-\evs_{X|a}^{*}\right\|^{2}\\
\delta_{anticausal}&=\left\|\evs_{A}-\evs_{A}^{*}\right\|^{2}+\sum_{a}\left\|\evs_{Y|a}-\evs_{Y|a}^{*}\right\|^{2}+\sum_{a,y}\left\|\evs_{X|a,y}-\evs_{X|a,y}^{*}\right\|^{2} 
\end{align*}

Using the previous conclusion in Section~\ref{sec:proof for p1},\\
$\sum_{a,y}\left\|\evs_{X|a,y}-\evs_{X|y}^{*}\right\|^{2}=K\sum_{a}\left\|\evs_{X|a}-\evs_{X|a}^{*}\right\|^{2} \ge \sum_{a}\left\|\evs_{X|a}-\evs_{X|a}^{*}\right\|^{2}$, we can get the relation: $\delta_{anticausal}\ge \delta_{causal}$.

\subsection{Proof for Proposition~\ref{pro:effect}}
\label{sec:proof for p2}
\begin{align*}
    \delta_{causal}&=\sum_{a,x}\left\|\evs_{Y|x,a}-\evs_{Y}^{*}\right\|^{2}\\
&=\sum_{a,x,y}(\evs_{y|x,a}-\theta_1(a,y))^{2}+\sum_{a,x,y}(\theta_1(a,y)-\evs_{y}^{*})^{2}\\
\delta_{anticausal}&=\sum_{a}\left\|\evs_{Y|a}-\evs_{Y}^{*}\right\|^{2}+\sum_{a,y}\left\|\evs_{X|a,y}-\evs_{X|a,y}^{*}\right\|^{2}\\ 
=\sum_{a}\left\|\evs_{Y|a}-\evs_{Y}^{*}\right\|^{2}&+\sum_{a,x,y}(\evs_{x|a,y}-\theta_2(a,x))^{2}+\sum_{a,x,y}(\theta_2(a,x)-\evs_{x})^{2}\\
\delta_{causal}-\delta_{anticausal}
&= (K-1) \|\vs^*_Y - \vc\|^2-(K-1)R^2,
\end{align*}
where $R^2= \frac{ (K-1) \|\vc\|^2 
     - K\|\bm\theta_1\|^2  + \|\vs_{Y|A}\|^2 
    + K\|\bm\theta_2 - \vs_X\|^2}{K-1}$,$\vc = \frac{K\bm\theta_1 - \vs_{Y|A} }{K-1}$
and $\theta_1 = \frac{(K-1)\vc + \vs_{Y|A}}{K}$. When $ \|\vs^*_Y - \vc\|^2<R^2$, $\delta_{causal}\ge\delta_{anticausal}$.

\onecolumn
\begin{multicols}{2}
\bibliographystyle{ACM-Reference-Format}
\bibliography{refer}
\end{multicols}


\end{document}